\documentclass[journal]{IEEEtran}
\usepackage[utf8]{inputenc} 
\usepackage[T1]{fontenc}    

\usepackage{graphicx}
\usepackage{tikz}
\usepackage{hyperref}
\usepackage{amsmath,amsfonts,amsthm,amssymb,dsfont}
\usepackage{stmaryrd}
\usepackage{subfig}
\usepackage{cite}

\newtheorem{definition}{Definition}

\newcommand\il[1]{\langle #1 \rangle}

\def\Z{\mathbb{Z}}
\def\R{\mathbb{R}}
\def\E{\mathcal{E}}
\def\D{\mathcal{D}}

\def\D{\mathcal{D}}
\def\E{\mathcal{E}}
\def\Z{\mathcal{Z}}
\def\LL{\mathcal{L}}

\newcommand{\MSE}{\mathrm{MSE}}
\newcommand\our{SeGMA}

\def\1F1{\mbox{$_{1}{F}_{\!1}$}}

\usepackage{color}

\usepackage{todonotes}

\newcommand{\ms}[1]{\textcolor{black}{#1}}
\newcommand{\mw}[1]{\textcolor{black}{#1}} 
\newcommand{\newmw}[1]{\textcolor{black}{#1}}

\begin{document}
\title{\our{}: Semi-Supervised Gaussian Mixture Auto-Encoder}
\author{Marek \'Smieja, Maciej Wo\l{}czyk, Jacek Tabor, and Bernhard C. Geiger,~\IEEEmembership{Senior Member,~IEEE}
\thanks{The work of Marek \'Smieja was supported by the National Science Centre (Poland) grant no. 2016/21/D/ST6/00980. 
Maciej Wo\l{}czyk carried out this work within the research project "Bio-inspired artificial neural
networks" (grant no. POIR.04.04.00-00-14DE/18-00) within the Team-Net program of the Foundation
for Polish Science co-financed by the European Union under the European Regional Development
Fund. The work of Jacek Tabor was supported by the National Science Centre (Poland) grant no. 2017/25/B/ST6/01271.
The work of Bernhard C. Geiger was supported by the COMET program within the K2 Center “Integrated Computational Material, Process and Product Engineering (IC-MPPE)” (Project No 859480). This program is supported by the Austrian Federal Ministries for Transport, Innovation and Technology (BMVIT) and for Digital and Economic Affairs (BMDW), represented by the Austrian Research Promotion Agency (FFG), and the
federal states of Styria, Upper Austria and Tyrol.
The Know-Center is funded within the Austrian COMET Program - Competence Centers for Excellent Technologies - under the auspices of the Austrian Federal Ministry of Transport, Innovation and Technology, the Austrian Federal Ministry of Digital and Economic Affairs, and by the State of Styria. COMET is managed by the Austrian Research Promotion Agency FFG.}%
\thanks{Marek \'Smieja, Maciej Wo\l{}czyk, and Jacek Tabor are with the Faculty of Mathematics and Computer Science, Jagiellonian University, \L{}ojasiewicza 6, 30-348 Krak\'ow, Poland (\mbox{e-mail:} \mbox{marek.smieja@uj.edu.pl}).}%
\thanks{Bernhard C. Geiger is with Know-Center GmbH, 8010, Graz, Austria (\mbox{e-mail:} \mbox{geiger@ieee.org}).}
}
%
%
%
\maketitle              
\begin{abstract}
We propose a semi-supervised generative model, \our{}, which learns a joint probability distribution of data and their classes and which is implemented in a typical Wasserstein auto-encoder framework. We choose a mixture of Gaussians as a target distribution in latent space, which provides a natural splitting of data into clusters. To connect Gaussian components with correct classes, we use a small amount of labeled data and a Gaussian classifier induced by the target distribution. \our{} is optimized efficiently due to the use of Cramer-Wold distance as a maximum mean discrepancy penalty, which yields a closed-form expression for a mixture of spherical Gaussian components and thus obviates the need of sampling. While \our{} preserves all properties of its semi-supervised predecessors and achieves at least as good generative performance on standard benchmark data sets, it presents additional features: (a) interpolation between any pair of points in the latent space produces realistically-looking samples; (b) combining the interpolation property with disentangling of class and style information, \our{} is able to perform \emph{continuous} style transfer from one class to another; (c) it is possible to change the intensity of class characteristics in a data point by moving the latent representation of the data point away from specific Gaussian components.
\end{abstract}

\section{Introduction}

Deep generative models, such as the variational auto-encoder (VAE, \cite{kingma2013auto}), Wasserstein auto-encoder (WAE, \cite{tolstikhin2017wasserstein}), deep Boltzmann machines~\cite{salakhutdinov2009dbm}, or generative adversarial networks (GAN, \cite{goodfellow2014generative}) play an important role in deep learning. They are capable of learning a probability distribution of complex high-dimensional data, which can be used to generate new examples. Their applications include image inpainting \cite{yeh2017semantic}, 3D object reconstruction \cite{fan2017point}, sentence paraphrasing \cite{guu2018generating}, and many more~\cite{zhu2016generative, zhu2017unpaired, shrivastava2017learning}.

In more specialized applications where data is composed of classes, a natural question is how to build a generative model that is able to generate examples from particular classes.
Since, in practice, the access to labeled data is limited, it is often not possible to train an individual generative model for each class using classical techniques. A step towards solving this problem was the introduction of semi-supervised versions of VAE models \cite{kingma2014semi, maaloe2016auxiliary}. Technically, these models learn an additional  categorical, i.e., discrete latent variable for encoding class labels by making use of a small number of labeled samples together with a large collection of unlabeled data. While they still describe the whole data distribution as their unsupervised predecessors, they are also capable of generating samples from specific classes. These samples are produced by transforming the discrete variable describing the class together with a latent code generated from a continuous target distribution using a decoder network.

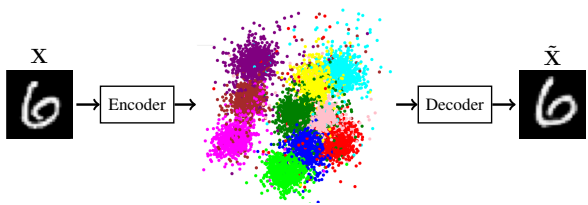
\begin{figure}[t]
    \centering
    \scalebox{0.65}{
    \begin{tikzpicture}
        \node[inner sep=2pt] (originals) at (0, 0)
        {\includegraphics[width=.075\textwidth]{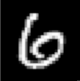}};
        \node at (0, 1.) {\large $\mathbf{X}$};
        \node (encoder) at (2, 0) [draw, thick, minimum width=1.5cm, minimum height=0.75cm] {Encoder};
        \draw[->, line width=0.5mm] (originals.east) -- (encoder.west);
        \node[inner sep=0pt] (mixture) at (5.25, 0)
        {\includegraphics[width=.225\textwidth]{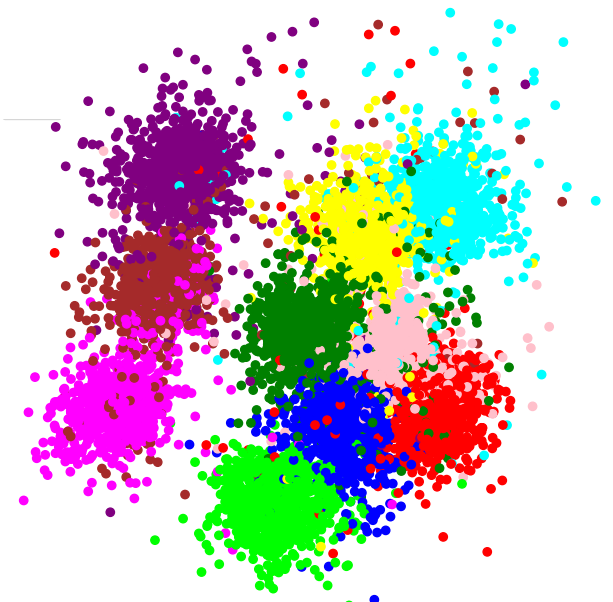}};
        \draw[->, line width=0.5mm] (encoder.east) -- (mixture.west);
        \node (decoder) at (8.5, 0) [draw, thick, minimum width=1.5cm, minimum height=0.75cm] {Decoder};
        \draw[->, line width=0.5mm] (mixture.east) -- (decoder.west);
        \node[inner sep=2pt] (rec) at (10.5, 0)
        {\includegraphics[width=.075\textwidth]{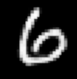}};
        \node at (10.5, 1.) {\large $\mathbf{\tilde{X}}$};
        \draw[->, line width=0.5mm] (decoder.east) -- (rec.west);
    \end{tikzpicture}
    }
    \caption{\our{} fits encoded data to a Gaussian mixture. Components are connected with classes using a small number of labeled data.}
    \label{fig:model_illustration}
\end{figure}

In this paper, we present an alternative approach for constructing generative models capable of creating class-specific samples. Instead of introducing a classification network with an additional discrete latent variable representing the class, we use a classical mixture of Gaussians to provide a natural disentanglement of class and style\footnote{\ms{Following \cite{kingma2014semi}, we decompose the information about each sample into two parts. The first factor includes the information needed to identify a class, while the second one accounts for variations within a class. This latter component is commonly referred to as \emph{style}.}} information in a single latent space of the auto-encoder. Using a small labeled data set, classes are assigned to components of this mixture of Gaussians by minimizing the cross-entropy loss induced by the class posterior distribution of a simple Gaussian classifier. The resulting mixture describes the distribution of the whole data, and representatives of individual classes are generated by sampling from its components, see Fig.~\ref{fig:model_illustration} for illustration.

\begin{figure}[t]
    \centering
    \includegraphics[width=0.5\textwidth]{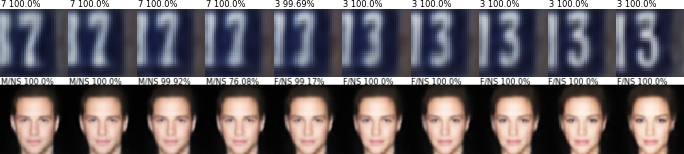}
    \caption{Continuous style transfer obtained by \our{}.}
    \label{fig:intro-style}
\end{figure}

Our model, which we will call semi-supervised Gaussian mixture auto-encoder (\our{}), is an adaptation of the Cramer-Wold auto-encoder (CWAE)~\cite{tabor2018cramer}, which is an instance of WAE models with maximum mean discrepancy (MMD) penalty~\cite{tolstikhin2017wasserstein}. The use of CWAE yields a closed-form expression for the MMD penalty for spherical Gaussians, which eliminates the need of sampling and propagating gradients through discrete variables which is not possible in a straightforward manner~\cite[Sec.~1.1]{rolfe2017dvae}. Experimental results on MNIST, SVHN, and CelebA data sets show that \our{} obtains comparable or even better generative performance than corresponding VAE models. Moreover, it has several advantages, which cannot be obtained using previous models:
\begin{itemize}
    \item Due to the use of a single continuous latent space in which the class information is encoded implicitly, we can interpolate between any two data points as in classical unsupervised generative models. In consequence, we are able to generate samples which mix properties of several classes. This property is difficult to obtain for semi-supervised VAE models, because they are trained directly on discrete class variables for which intermediate states are not necessarily meaningful.
    \item Despite lacking an architectural separation between the discrete class label and the continuous latent code, we demonstrate that \our{} is able to disentangle the class information of a data point from its style. In contrast to semi-supervised VAE models, in which this disentanglement is explicit, \our{} allows to perform continuous style transfer. In other words, we can change the class variable smoothly, while preserving the style of examples. This is a consequence of interpolation property, see Fig.~\ref{fig:intro-style}. While continuous style transfer is performed by moving a latent representation from the original Gaussian component \emph{towards} the component of the target class, the intensity of the style w.r.t.\ certain class-discriminating characteristics can be increased by moving the latent representation \emph{away from} specific Gaussian components.
    \item \our{} is easy to implement in an existing WAE-MMD framework and is robust to the choice of hyperparameters (see Section~\ref{sec:sensitivity}). We achieve better generative performance than related VAE approaches despite using simpler neural networks and significantly lower-dimensional latent space -- e.g., we use a classical encoder-decoder network with a 20-dimensional latent space for SVHN, while the method proposed by~\cite{maaloe2016auxiliary} consists of six networks and 300 dimensions for encoding latent codes. \our{} is trained faster and in a more stable way due to the use of Cramer-Wold distance which eliminates the need of sampling. Finally, since we use a classical auto-encoder architecture, it is easier to apply convolutional networks than in semi-supervised VAE models, where continuous and discrete inputs have to be gathered together.
\end{itemize}

\ms{To give a concrete real-life example where the proposed model is useful, let us consider a graphical application for modifying faces. More precisely, the application takes a face image and allows to modify its selected features, such as smile, type of beard, age, etc. Clearly, it should be possible to change the beard length, choose a gentle or wide smile as well as let the face age. We emphasize that it is desirable to be able to change these features smoothly, rather than having to choose among few discrete options. Our model satisfies these requirements by interpolating between features and changing the intensity of labels as can be seen in Fig.~\ref{fig:intro-style}. For more details, see Sections \ref{sec:interpolation} and \ref{sec:intensity}.}

The remainder of this work is organized as follows: We discuss the similarities between our approach and the related works in semi-supervised and unsupervised generative models in Section~\ref{sec:related}. The WAE and the Cramer-Wold distance are revised in Section~\ref{sec:preliminaries}. We complete the theoretical framework of \our{} in Section~\ref{sec:semi-supervised}, where we include a regularization term based on a small labeled data set. For the specific choice of a mixture of Gaussians as a target distribution in latent space, we instantiate our theory in Section~\ref{sec:practical} and briefly discuss optimization issues. We conclude by showing experimental results and comparisons with state-of-the-art semi-supervised VAE models in Section~\ref{sec:experiments}.

\section{Related work}\label{sec:related}

There are a few works on unsupervised deep clustering, which use a mixture of Gaussians as a target distribution in latent space.
The authors of~\cite{jiang2017vade} propose a generative model based on the VAE in which both the encoder and decoder are stochastic.\footnote{\ms{We follow the literature and call a neural network (encoder/decoder) stochastic if the network parameterizes a probability distribution, i.e., if the network output is sampled from a distribution parameterized by the network parameters and the input. We call a neural network deterministic if the network parameterizes a function, i.e., if the network output is a function of the input parameterized by the network parameters.}} 
They train their model by maximizing the evidence lower bound (ELBO) over the parameters of the Gaussian mixture model and the parameters of the encoder and decoder. 
A very similar model was recently introduced in~\cite{ugur2019gmmvib}, with the main difference that the decoder is deterministic and that, instead of maximizing the ELBO, a variational bound on the information bottleneck functional is optimized, cf.~\cite{alemi2017dvib}, which is simultaneously an upper bound to the ELBO proposed in~\cite{jiang2017vade} for certain parameter settings, cf.~\cite[Remark~1]{ugur2019gmmvib}. 


Despite being unsupervised, both~\cite{jiang2017vade} and~\cite{ugur2019gmmvib} report that the obtained clusters agree very well with the class labels. However,~\cite{jiang2017vade,ugur2019gmmvib}, and also~\cite{alemi2017dvib,gaujac2018gaussian}, evaluated the generative performance of their models only on MNIST. For more complicated datasets, an agreement between clusters and classes need not occur. Indeed,~\cite{dilokthanakul2016gmmvae}, proposed a Gaussian mixture model in which the means and variances are learned by training the parameters of an additional neural network with a standard Gaussian distribution at its input. The authors show that the clusters coincide with the individual classes for MNIST, but tend to group images that are visually similar for SVHN, respectively; cf.~\cite[Figs.~5 \& 6]{dilokthanakul2016gmmvae}.


Another approach for training deep generative models relies on using WAE \cite{tolstikhin2017wasserstein}. The authors of \cite{kolouri2018sliced} show that in some cases the optimization of a classical shallow Gaussian mixture model using sliced p-Wasserstein distance \cite{kolouri2017optimal} leads to better parameters than applying the EM algorithm. Instead of a joint optimization of Wasserstein distance with neural network weights, they fit a Gaussian mixture model on a fixed embedding of data.
Reference~\cite{gaujac2018gaussian} explains the difficulties in training VAE models with discrete latent variables \cite{lawson2018learning, jang2016categorical, maddison2016concrete}. As a remedy they propose a WAE-MMD model with a mixture of Gaussians as a target distribution. This fully unsupervised model uses discrete and continuous latent variables and is optimized by sampling from a target distribution.

The semi-supervised deep generative models proposed in the literature differ substantially from the unsupervised generative models discussed so far. The most important difference is that they have two explicit latent variables, a discrete one accounting for the class and a continuous one, commonly attributed to represent the style of the data point. Thus, these methods explicitly consider the discrete variable also during inference, e.g., by training a classifier network on the latent space in addition to an encoder network. For example,~\cite{kingma2014semi} proposed a generative model parametrized by a stochastic neural network taking the latent class and style variables as inputs. Similarly, encoder and classifier -- computing the style and class variables, respectively -- are implemented as stochastic neural networks. All model parameters are optimized by maximizing the ELBO over the unlabeled and labeled parts of the data set; for the former, the class variable is supplied via posterior inference using the classifier network. The authors of~\cite{maaloe2016auxiliary} improved upon~\cite{kingma2014semi} by adding a continuous auxiliary latent variable, thereby increasing the capacity of the model and improving the variational bound. By using neural networks for class inference, both~\cite{kingma2014semi} and~\cite{maaloe2016auxiliary} achieve good classification and generative performance with only a small labeled data set. The paper \cite{maaloe2017cagem} follows a different approach and modifies the VAE model so that one of the continuous latent variables represents a clustering structure of the data. The authors of \cite{dai2017good} build their semi-supervised model on GANs, but they focus directly on a classification task. They show that a generator that models the data distribution perfectly neither leads to a better classifier nor utilizes unlabeled data effectively. To achieve good classification performance, they thus propose a generator that does not produce realistic samples.

A relevant example from supervised learning is~\cite{Moyer_InvariantDVIB}. Combining the loss function of the VAE with a regularizer that rewards latent codes that are independent of the class variable, the authors obtain a generative model that creates MNIST images from the latent code (thus representing only style) and a discrete class variable that they feed as an additional input.

On the one hand, our setting is conceptually similar to generative deep clustering approaches which use mixtures of Gaussians as target distributions. Most of these works use a single continuous latent space (except the WAE-MMD approach \cite{gaujac2018gaussian}, which uses two latent variables), but require sampling from some distribution during training. In contrast, the use of the Cramer-Wold distance in \our{} allows fitting the encoded data to the target mixture using a closed-form expression. Furthermore, our focus is on the generative properties rather than on clustering. In comparison,~\cite{jiang2017vade, gaujac2018gaussian} generated samples only for MNIST,~\cite{ugur2019gmmvib} did not generate samples at all, and none of the mentioned works evaluate the quality of their samples using quantitative methods. 

On the other hand, our work draws concepts from semi-supervised VAEs, which are comprised of (at least) two latent spaces, with one of them being discrete. The main difference to semi-supervised VAEs is that we propose using a single continuous latent space that allows us not only to transfer style to different classes, but also to interpolate between data points from different classes. Furthermore, we eliminate the need for an auxiliary classifier network by simply computing the loss of a Gaussian classifier based on a small labeled data set. Finally, our cost function is not based on variational principles, but is motivated by WAEs.


\section{Preliminaries: Wasserstein Auto-Encoder and Cramer-Wold Distance} \label{sec:preliminaries}

Our approach is based on the WAE with a MMD penalty~\cite{tolstikhin2017wasserstein}. As opposed to the VAE~\cite{kingma2013auto}, WAE allows using deterministic, i.e., non-random, encoders mapping the input to a latent space via a function. To eliminate the need for sampling during training and to improve the model stability, we implement MMD using Cramer-Wold distance \cite{tabor2018cramer}.

Let $X = (x_i)_{i=1}^n \subset \R^N$ be a data set. A non-random auto-encoder consists of a deterministic encoder $\E{:}\ \R^N \to \Z$ and a deterministic decoder $\D{:}\ \Z \to \R^N$, where $\Z = \R^D$ is a latent space (usually $D < N$). The encoder thus provides a lower-dimensional representation of $X$ that the decoder shall be able to reconstruct with as small an error as possible. This goal is achieved by minimizing the mean squared error between input and output of the auto-encoder:
\begin{equation}\label{eq:MSE}
    \MSE(X; \E,\D) = \frac{1}{n} \sum_{i=1}^n \|x_i - \D(\E(x_i))\|^2,
\end{equation}
where $\|\cdot\|$ denotes the Euclidean norm.

To make the auto-encoder generative, the reconstruction error is complemented by a regularization term that encourages the distribution of the low-dimensional representation $\E X$ to be similar to a target distribution $P_Z$. Specifically, for a WAE with MMD penalty, the regularization term is chosen as the norm in some reproducing kernel Hilbert space $\mathcal{H}_k$, i.e., 
\begin{multline} \label{eq:mmd}
    d_{\mathrm{MMD}}(\E X, P_Z) \\= \Big\Vert \int_\mathcal{Z} k(z,\cdot) \mathrm{d}P_{\E X}(z) - \int_\mathcal{Z} k(z,\cdot) \mathrm{d}P_Z(z) \Big\Vert_{\mathcal{H}_k},
\end{multline}
where $P_{\E X}$ is the distribution of $\E X$ and where $k{:}\ \mathcal{Z}\times\mathcal{Z}\to \mathbb{R}$ is a positive definite reproducing kernel. If the distribution of $\E X$ is similar to $P_Z$ and if $\E$ and $\D$ are such that $\MSE(X; \E,\D)$ is small, then samples from $X$ can be generated by sampling from $P_Z$ and passing these samples to the decoder $\D$~\cite[Remark~4.1]{tabor2018cramer}. 

During the training process, the comparison of the latent data representations $\E X$ with a target distribution $P_Z$ is performed by sampling from $P_Z$ and approximating \eqref{eq:mmd} by averaging the kernel $k$ evaluated at all pairs of samples \cite[Algorithm 2]{tolstikhin2017wasserstein}. Such a procedure is necessary even for a simple target distribution such as the standard Gaussian $N(0,I)$. To eliminate the need for sampling, MMD can be implemented using the Cramer-Wold distance, which has a closed-form expression for spherical Gaussian distributions $N(\mu,\sigma^2 I)$. 
This approach results in Cramer-Wold auto-encoder (CWAE) \cite{tabor2018cramer}, which is an instance of WAE-MMD with inverse multiquadratics kernel \cite[page 9]{tolstikhin2017wasserstein}.

The Cramer-Wold distance $d^2_\gamma$, where $\gamma > 0$ is a regularization parameter, is induced by a corresponding scalar product $\il{\cdot,\cdot}_\gamma$, which for two spherical, $D$-dimensional Gaussians $N(a,\alpha I), N(b,\beta I)$ is given by \cite[eq. 11]{tabor2018cramer}:
\begin{multline}\label{eq:cwprod}
\il{N(a,\alpha I), N(b, \beta I)}_\gamma 
\\ = \frac{1}{\sqrt{2\pi(\alpha + \beta + 2 \gamma)}} \phi_D \left(\frac{\|a-b\|^2}{2(\alpha + \beta + 2 \gamma)} \right),
\end{multline}
where $\phi_D(s) \approx (1+\tfrac{4s}{2D-3})^{-1/2}$, for $D \geq 10$.

\ms{To evaluate~\eqref{eq:cwprod} for a sample $Z = (z_i)_{i=1}^m$, we represent it as the uniform probability measure on this sample $\frac{1}{m} \sum_{i=1}^m \delta_{z_i}$, where $\delta_{z}$ is an atom Dirac measure condensed at $z$. Note that a Dirac measure $\delta_z$ can be interpreted as a degenerate Gaussian density\footnote{A degenerate Gaussian density has a singular (non-invertible) covariance matrix, see \cite{rao1973linear}.} with  mean $z$ and the zero matrix as covariance matrix\footnote{Such a Gaussian density is spanned on 0-dimensional linear subspace.}. Consequently, a sample $Z = (z_i)_{i=1}^m$ is represented as the uniform mixture of degenerate Gaussians with zero covariance, i.e. $Z = \frac{1}{m} \sum_{i=1}^m N(z_i,0)$.} If we put $P_Z = N(0,I)$, then the Cramer-Wold distance
between this target distribution and the sample $Z$ equals:
\begin{align}\label{eq:CW_Gauss_Data}
&d^2_\gamma(Z,P_Z)= \il{Z-P_Z, Z-P_Z}_\gamma \notag\\
 &= \frac{1}{m^2}\sum_{i=1}^m \sum_{j=1}^m \il{N(z_i,0), N(z_j,0)}_\gamma \notag\\
&\quad\quad - \frac{2}{m} \sum_{i=1}^m \il{N(z_i,0), N(0,I)}_\gamma
+ \il{N(0, I), N(0,I)}_\gamma
\end{align}
where $\il{\cdot,\cdot}_\gamma$ is given by \eqref{eq:cwprod}. The parameter $\gamma$ can be seen as a kernel bandwidth and is commonly selected using Silverman's rule of thumb for Gaussian densities. 



\section{Theoretical Model: Semi-Supervised Learning for Improved Class Representation}\label{sec:semi-supervised}

Classical WAE models allow us to describe the distribution of the data and to generate samples from it. In practice, data can come from several distinct classes. In such a case, we want to model a distribution of the data as a whole as well as to be able to generate samples from a given class. 

To this end, we introduce a joint distribution $P_{Z,Y}$, which additionally encodes the latent class variable $Y$. On the one hand, marginalization over the class variable $Y$ yields the target distribution $P_Z$ to which the WAE fits the latent representation. On the other hand, to generate samples from a given class $y$, we fix this class in the distribution $P_{Z,Y}$, sample from the resulting conditional distribution $P_{Z|Y=y}$, and pass the obtained latent codes through the decoder $\D$. Thus, $P_{Z,Y}$ should be parametrized such that both:
\begin{itemize}
    \item samples from its marginal $P_Z$ have high quality,
    \item samples from $P_{Z|Y=y}$ are representatives of class $y$.
\end{itemize}
In practical cases, $P_Z$ could be a multi-modal distribution, with each mode corresponding to one class.

To parametrize $P_{Z,Y}$ (e.g., to assign the modes of the distribution $P_Z$ to individual classes), we follow a typical semi-supervised learning scenario and use a small number of labeled examples \cite{wang2013safety}. It is assumed that we have access to a labeled data set $(X_L, Y_L) = (x_i,y_i)_{i=n+1}^m$, where $X_L \subset \R^N$ and where $y_i \in \{1,\ldots,K\}$ denotes the class label. To incorporate knowledge coming from labeled examples in building our generative model, we use a classifier. More precisely, let $P_{Y|Z}$ be the class posterior probability and let $\hat{P}_{Y_L,\E X_L}=\frac{1}{m-n}\sum_{i=n+1}^{m}\delta[(y,z)=(y_i,\E x_i)]$ be the empirical joint distribution of the labeled encoded data set, where $\delta[\cdot]$ is the Kronecker delta. We complement the WAE objective with the cross-entropy between the true class label and the class label inferred from $P_{Y|Z}$:
\begin{align}\label{eq:logloss}
    &\LL(\hat{P}_{Y_L|\E X_L}\Vert P_{Y|Z})\notag\\
    & = \frac{1}{m-n} \sum_{i=n+1}^m -\log P_{Y|Z}(y_i|\E(x_i))\notag\\
    & = \frac{1}{m-n} \sum_{i=n+1}^m -\log \frac{P_{Y}(y_i) P_{Z|Y=y_i}(\E(x_i))}{P_{Z}(\E(x_i))}
\end{align}
This suggests that a good generative model in the proposed semi-supervised setting consists of an encoder $\E$ and a decoder $\D$ such that
\begin{equation}\label{eq:general_cost}
    \MSE(\hat{X}; \E,\D) + \alpha d_{\mathrm{MMD}}(\E \hat{X}, P_Z)+\beta\LL(\hat{P}_{Y_L|\E X_L}\Vert P_{Y|Z})
\end{equation}
is minimized, where $\hat{X} = X \cup X_L$ and $\alpha, \beta$ are trade-off parameters.

\section{Practical Implementation}\label{sec:practical}

In this section, we present our choice for the distribution $P_{Z,Y}$, instantiate the cost function~\eqref{eq:general_cost} for this choice and discuss its optimization.

In order to keep the target distribution simple, we choose $P_{Z,Y}$ to be a Gaussian mixture model. In this case, $P_Y$ is a categorical distribution from $K$ categories with prior probabilities $p_1,\ldots,p_K$, while $P_{Z|Y=k}$ is a multivariate Gaussian $N(\mu_k,\Sigma_k)$, which describes the distribution of class $k$ in latent space. The marginalization of $P_{Z,Y}$ over classes $Y$ gives a target distribution $P_Z$ which is a mixture of Gaussians with density $P_Z=\sum_{k=1}^K p_k N(\mu_k,\Sigma_k)$. 

On the one hand, choosing $P_{Z,Y}$ to be a Gaussian mixture model makes it easy to generate latent codes for a given class $k$ by sampling from the conditional density $P_{Z|Y=k}=N(\mu_k,\Sigma_k)$. At the same time, the posterior probability used in the semi-supervised setting simplifies to: 
\begin{equation} \label{eq:postG}
    P_{Y|Z}(k|z) = \frac{p_k N(\mu_k, \Sigma_k)(z)}{\sum_{l=1}^K p_l N(\mu_l, \Sigma_l)(z)}.
\end{equation}
Using such a simple Gaussian classifier in the feature space directly may fail; we show in our experiments that its embedding in the latent space yields satisfactory performance.

On the other hand, there appears the question how to compute $d_{\mathrm{MMD}}(\E X,P_Z)$ for the mixture of Gaussians and how to train a network minimizing the WAE objective in this case. The simplest approach relies on sampling from the mixture density $P_Z$ and next computing the MMD distance between these samples and encoded data. The implementation of this step requires the reparameterization trick to back-propagate the gradients. However, since we have $K$ Gaussians, the number of samples should probably be $K$ times larger than for a single Gaussian to approximate this this distribution well enough, which might negatively influence the training time. 

Our choice to implement the MMD penalty with the Cramer-Wold distance \cite{tabor2018cramer} resolves this issue since it obviates the need for sampling, effectively decreasing training time and improving model stability \ms{(see \cite[Section 7.4]{tabor2018cramer} for a comparison between CWAE, WAE and VAE)}. Combining~\eqref{eq:cwprod} and~\eqref{eq:CW_Gauss_Data}, the Cramer-Wold distance between a sample $Z = (z_i)_{i=1}^m$ and a mixture of spherical Gaussians $P=\sum_{k=1}^K p_k N(\mu_k,\sigma^2_k I)$ can be computed in closed form:
\begin{align}\label{eq:CW-GMM}
&d^2_\gamma(Z,P)=\frac{1}{m^2\sqrt{2\pi(2 \gamma)}}\sum_{i=1}^m\sum_{j=1}^m\phi_D\left(\frac{\|z_i-z_j\|^2}{4\gamma}\right)\notag\\&\quad -\sum_{i=1}^m\sum_{k=1}^K \frac{2p_k}{m\sqrt{2\pi(\sigma^2_k+2\gamma)}}\phi_D\left(\frac{\|z_i-\mu_k\|^2}{2(\sigma^2_k+2\gamma)}\right)\notag\\
 &\quad+\sum_{k=1}^K\sum_{l=1}^K\frac{p_kp_l}{\sqrt{2\pi(\sigma^2_k + \sigma^2_l + 2\gamma)}} \phi_D\left(\frac{\|\mu_k-\mu_l\|^2}{2(\sigma^2_k + \sigma^2_l+2\gamma)}\right).
\end{align}

To use the CWAE with the target distribution $P_Z$ being a mixture of Gaussians, it is sufficient to replace $d_{\mathrm{MMD}}$ in~\eqref{eq:general_cost} by~\eqref{eq:CW-GMM}. We further take the logarithm of the squared Cramer-Wold distance to improve the balance between this term and the MSE~\cite[Footnote 2]{tabor2018cramer}. Taking all this together, we define the \our{} objective function.
\begin{definition}\label{def:cost}
Let $P_Z = \sum_{k=1}^K p_k N(\mu_k,I)$ be a target distribution and let $\hat{X} = X \cup X_L$ be composed of unlabeled and labeled data, where $Y_L$ denotes labels of $X_L$. The aim of \our{} is to find minimizers of the following optimization problem:
 \begin{multline}\label{eq:cost}
    \min_{\E,\D,\{\mu_k\},\{p_k\}}\quad \MSE(\hat{X}; \E,\D) \\+ \alpha \log d_\gamma^2(\E \hat{X}, P_Z)+\beta\LL(\hat{P}_{Y_L|\E X_L}\Vert P_{Y|Z}),
\end{multline}
where $\alpha, \beta$ are trade-off parameters and where $P_{Y|Z}$ and $d_\gamma^2$ are given in~\eqref{eq:postG} and \eqref{eq:CW-GMM}, respectively.
\end{definition}

As it can be seen in this definition, we choose to model each class with a single Gaussian component with identity covariance matrix. One may argue that this choice is very restrictive, limiting the expressiveness of our model. In practice, however, encoder and decoder are implemented by deep neural networks. Thus, while multiple components with trainable variances per class may lead to a lower loss, our preliminary experiments indicate that the restrictions for $P_Z$ in Definition~\ref{def:cost} are unproblematic from a practical perspective, given that the encoder and decoder networks have sufficient capacity. At the same time, the capacity of the encoder and decoder needs to be restricted to ensure that data points from the same class are mapped to the same component in latent space.

\our{} is trained similar as CWAE, with a slight modification caused by the use of labeled data. Every mini-batch, which is composed of labeled and unlabeled data, is processed by the encoder and decoder. It is common practice that half of data in a mini-batch is labeled\footnote{If the number of labeled data is too low to achieve this, we can either decrease the number of samples in a mini-batch, or use oversampling techniques to virtually increase the number of labeled data.} \cite{kingma2014semi}. If we have additional knowledge about class proportions, we use these proportions for selecting labeled examples in each mini-batch. The cross entropy $\LL$ is computed only for labeled examples, while $\MSE$ and $d_\gamma^2$ are evaluated for all data. During back-propagation, we update both weights of the encoder and decoder, as well as the means $\mu_k$ of Gaussians. Component masses $p_k$ are set to reflect the proportions of classes in the set of the known labels $Y_k$. While component masses can also be trained, our preliminary experiments show that this does not improve model performance significantly. Updating the means of Gaussians, however, allows us to better organize the class structure in the latent space. Namely, our model can place similar classes closer together, while significantly distinct ones can be put far apart. This property can be useful in analyzing the similarity between classes after successful training, which would not be possible using a separate discrete class variable.

\section{Experiments}\label{sec:experiments}

In this section, we present the results of our experiments. Our goal is to compare with similar methods on standard benchmark tasks as well as to present interesting properties of \our{} that are difficult to obtain using existing approaches. 

\subsection{Experimental setting}

For the evaluation, we chose three datasets of varying complexity, namely MNIST \cite{mnist}, SVHN \cite{svhn} and CelebA \cite{celeba}, which are commonly used for comparing generative models \cite{maaloe2016auxiliary, tolstikhin2017wasserstein}. For these datasets we randomly chose, respectively, 100, 1000 and 1000 labeled samples, while the rest of the total 60000, 604388 and 182637 samples remained unlabeled. Since CelebA labels consist of multiple tags (i.e. sex, presence of eyeglasses, presence of facial hair, etc.), we kept only the two most balanced tags ('smiling' and 'male') and created four classes: 'not male, not smiling', 'not male, smiling', 'male, not smiling' and 'male, smiling'.

We compare \our{} to previous semi-supervised generative models based on VAEs: M1+M2 \cite{kingma2014semi} and ADGM \cite{maaloe2016auxiliary}. \newmw{Additionally, we compare to Triple-GAN \cite{chongxuan2017triple}, which, however, does not allow for reconstructing examples. This rules out some of the experimental studies performed in this paper, e.g., interpolation, which explains why Triple-GAN does not appear in Sections~\ref{sec:interpolation} and~\ref{sec:intensity}. These methods were reimplemented based on an open repository}\footnote{https://github.com/wohlert/semi-supervised-pytorch}, the original authors' codes\footnote{https://github.com/dpkingma/nips14-ssl for M1+M2, https://github.com/larsmaaloee/auxiliary-deep-generative-models for ADGM \newmw{and https://github.com/zhenxuan00/triple-gan for Triple-GAN.}} and the description of these models in the papers\footnote{We we unable to use original authors codes directly, because they are based on a version of Theano that is no longer supported.}. We excluded methods that do not focus on generative properties (although they are based on generative models), e.g. \cite{dai2017good}, or which are not able to generate samples from a desired class, e.g. \cite{goodfellow2014generative}. To the best of our knowledge, these are the only deep models, which introduce partial supervision to learn probability distributions of specific classes (not only the whole data density). For a comparison between classical WAE, VAE and GAN, we refer the reader to \cite{tabor2018cramer} and \cite{tolstikhin2017wasserstein}.


Contrary to the pipelines presented in \cite{kingma2014semi, maaloe2016auxiliary}, we used no preprocessing on any of the datasets except for CelebA, where we followed \cite{tolstikhin2017wasserstein} and cropped and then resized images to a resolution of 140$\times$140 and 64$\times$64, respectively. \newmw{For experiments on MNIST and SVHN we kept the architectures from the original papers, i.e., fully connected architectures for M1+M2 and ADGM and convolutional networks in the case of Triple-GAN.} 
For a fair comparison \newmw{with VAE-based methods}, \our{} was also instantiated with fully connected networks -- we used ReLU activation functions and layer sizes $1024-1024-10-1024-1024-784$ and $786-786-786-786-786-20-786-786-786-786-786-3072$, respectively (i.e., the latent space has dimension $D=10$ and $D=20$ for MNIST and SVHN). \newmw{For CelebA, the architecture was based on the one from~\cite{tabor2018cramer} and then adapted to the unsupervised setting; the details can be found in the Appendix. We initialize the network parameters using the Glorot initialization~\cite{glorot2010understanding} and we initialize the means $\mu_k$ of the Gaussian components so that distance between each two means is $1$ at the start of the training.} To show that \our{} does not need much fine-tuning on the validation set (which is often impractical in semi-supervised settings with a small number of available labels), we used the same values of hyperparameters ($\alpha = 5$, $\beta=10$, $\gamma=(\frac{4}{3 N_c})^{\frac{2}{5}}$, where $N_c$ is the number of samples per class in a single batch, learning rate $=3\cdot10^{-4}$) for all experiments and adapted only the network architecture to each data set. The reader is referred to Section~\ref{sec:sensitivity}, where the sensitivity to some of the parameters is analyzed. We do not use any regularizers (weight decay, dropout, batch normalization), since we want to show that our model works in the simplest possible setting. Additional regularization might further improve its performance.


\begin{figure}[t!]
    \centering
    \subfloat[Triple-GAN]{
        \includegraphics[width=0.49\linewidth]{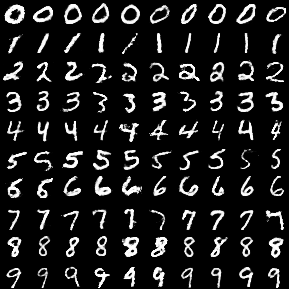}
    }
   \subfloat[M1+M2\label{fig:vae-mnist}]{
        \includegraphics[width=0.49\linewidth]{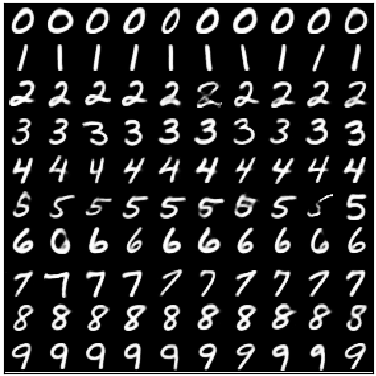}
    } \\
    \subfloat[ADGM]{
        \includegraphics[width=0.49\linewidth]{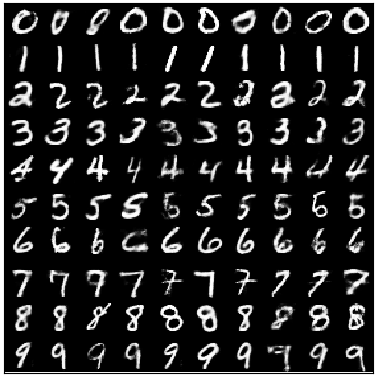}
    }  
     \subfloat[\our{}\label{fig:our-mnist}]{
        \includegraphics[width=0.49\linewidth]{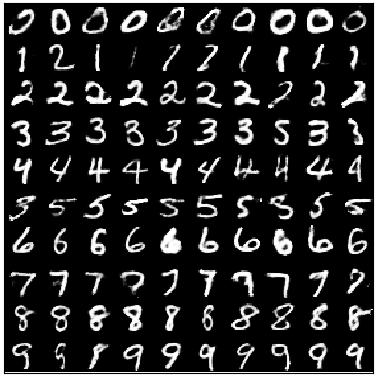}
    }
    \caption{Samples generated for MNIST.}
    \label{fig:mnist}
\end{figure}

\begin{figure}[t!]
    \centering
    \subfloat[Triple-GAN\label{fig:tgan-svhn}]{
        \includegraphics[width=0.49\linewidth]{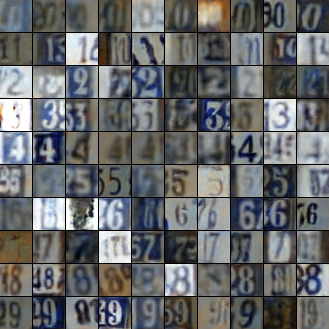}
    }
    \subfloat[M1+M2\label{fig:vae-svhn}]{
        \includegraphics[width=0.49\linewidth]{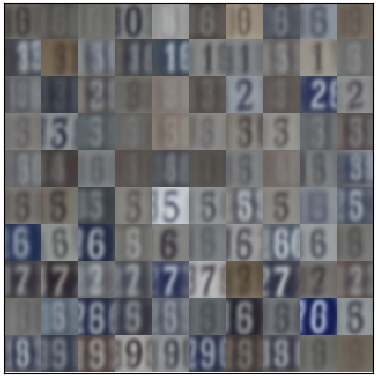}
    } \\
    \subfloat[ADGM\label{fig:adgm-svhn}]{
        \includegraphics[width=0.49\linewidth]{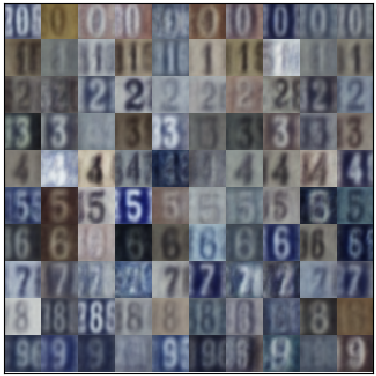}
    }
    \subfloat[\our{}\label{fig:our-svhn}]{
        \includegraphics[width=0.49\linewidth]{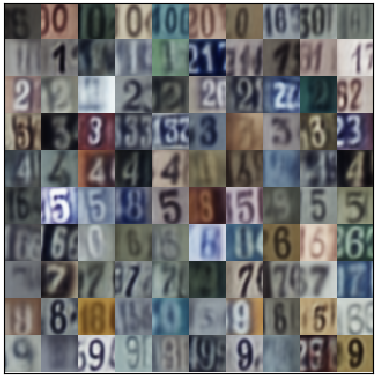}
    }
    \caption{Samples generated for SVHN.}
    \label{fig:svhn}
\end{figure}

\begin{figure}[t!]
    \centering
    \subfloat[M1+M2]{
        \includegraphics[width=0.98\linewidth]{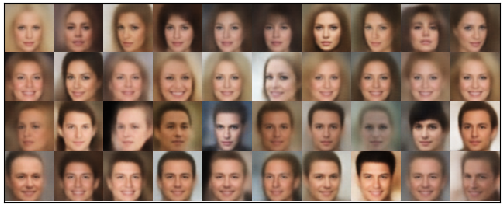}
    }
    \\
    \subfloat[ADGM]{
        \includegraphics[width=0.98\linewidth]{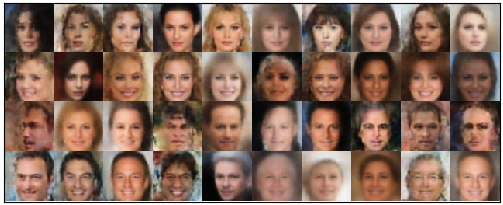}
    }
    \\
    \subfloat[\our{}]{
        \includegraphics[width=0.98\linewidth]{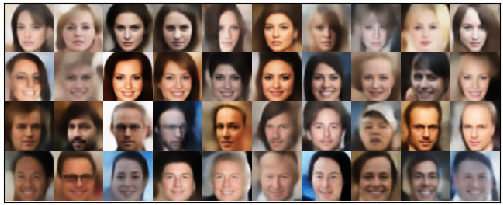}
    }
    \caption{Samples generated for CelebA.}
    \label{fig:celeba}
\end{figure}

\begin{table}[tp]
	\centering
	\caption{FID scores on various datasets (lower is better). The results in brackets for SVHN was obtained using convolutional architecture.}
	\setlength{\tabcolsep}{6pt}
	\label{tab:fid_scores}
	\begin{tabular}{llccc}
		\hline
		{\bf Variant }& {\bf Model} & {\bf MNIST} & {\bf SVHN} & {\bf CelebA} \\  
		\hline
		unsupervised & CWAE & 20.76 & 49.30 & 51.94 \\
		& \our{} & 17.79 & 61.11 & 51.13 \\ 
		\hline 
		& \newmw{Triple GAN} & 10.46 & 45.40 & -- \\
		& M1+M2 & 40.50 & 142.11 & 70.96 \\
		semi-supervised & ADGM & 14.50 & 55.54 & 51.96 \\
		& \our{} & 12.51 & 63.80 (50.10) & 50.54 \\
		\hline
	\end{tabular}
\end{table}

\begin{table}[tp]
	\centering
	\caption{Test errors on various datasets (lower is better). The results in brackets for SVHN was obtained using convolutional architecture.}
	\label{tab:accuracy}
	\begin{tabular}{lccc}
		\hline
		{\bf Model} & {\bf MNIST} & {\bf SVHN} & {\bf CelebA} \\ 
		\hline
		\newmw{Triple GAN} & 0.8\% & 5.8\% & -- \\
		M1+M2 & 17.0\% & 65.7\% & 44.5\%  \\
		ADGM & 4.1\%& 41.2\% & 41.7\% \\
		\our{} & 12.7\% & 32.8\% (22.4\%) & 24.6\% \\ 
		\hline
	\end{tabular}
\end{table}


\subsection{Generative performance} \label{sec:generativity}

First, examine the quality of generated samples and show that \our{} is capable of generating representative samples of a given class. For this purpose, we generate samples from each Gaussian component and pass these through the decoder. 

Randomly selected samples are presented in Figs.~\ref{fig:mnist}, \ref{fig:svhn} and \ref{fig:celeba} -- each row in the images corresponds to samples related to a separate Gaussian component for \our{}, and to samples determined by a separate discrete class variable in M1+M2, ADGM and Triple-GAN, respectively. It is evident that \our{} generates realistically-looking samples. The produced samples are diverse and as sharp as samples returned by the corresponding CWAE model, see \cite[Figure 2]{tabor2018cramer}. While ADGM generates samples of comparable quality for MNIST, the images produced for SVHN are less colorful. For CelebA, ADGM returns grainy images, which contain many anomalies. It may be caused by an encoder network, which has to combine convolutional layers applied to an image with dense layers applied to one-hot vector representing the class\footnote{The authors of ADGM did not examine their method in the case of convolutional neural networks.}. M1+M2 gives reasonable results only for MNIST while for other data sets, the images are blurry and there is not enough variance between different images which might follow from the mode collapse. We wish to mention that the results reported in \cite{kingma2014semi} displayed better quality for SVHN. We believe that this is due to specific preprocessing and careful hyperparameter selection, which is difficult to accomplish in a typical semi-supervised setting where the number of labeled data available for a validation stage is limited.
\newmw{Triple-GAN generates very realistically-looking and sharp samples, but at the same time produces anomalies (e.g. the third image in the seventh row of Fig.~\ref{fig:tgan-svhn}). We believe that the quality of Triple-GAN samples comes at the cost of extensive hyperparameter tuning and architecture search. In particular, we tried to use the architecture for Cifar-10 from the original work in order to run the method on CelebA dataset, but the method was not able to generate faces even after few hundred steps of training. For MNIST and SVHN, we used the architectures and hyperparameters from~\cite{chongxuan2017triple}; these were originally tuned on a held-out validation set with all labels \cite{chongxuan2017triple}, which is an impractical approach to semi-supervised learning~\cite{oliver2018realistic}. In Section \ref{sec:sensitivity} we show that such precise choice of hyperparameters is not required for \our{}.}

\begin{table}[tp]
	\centering
	\caption{FIDs of interpolations (lower is better). The results in brackets for SVHN was obtained using convolutional architecture.}
	\label{tab:interpolation_fids}
	\begin{tabular}{lccc}
	    \hline
		{\bf Model} & {\bf MNIST} & {\bf SVHN} & {\bf CelebA}\\ 
		\hline
		M1+M2 & 56.38 & 176.89 & 88.24\\
		ADGM & 21.75 & 76.66 & 63.60\\
		\our{} & 13.40 & 63.47 (50.01) & 48.14\\ 
		\hline
	\end{tabular}
\end{table}

\begin{figure*}[t]
\centering
    \subfloat[ADGM]{
        \includegraphics[width=.24\linewidth]{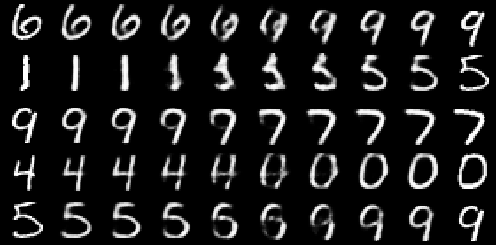}
    }
    \subfloat[\our{}]{
        \includegraphics[width=.24\linewidth]{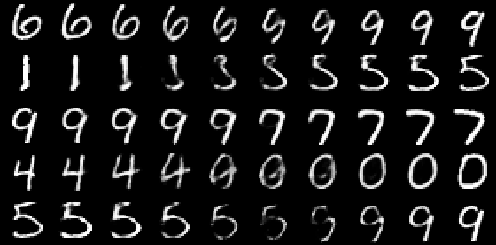}
    }
    \subfloat[ADGM]{
        \includegraphics[width=.24\linewidth]{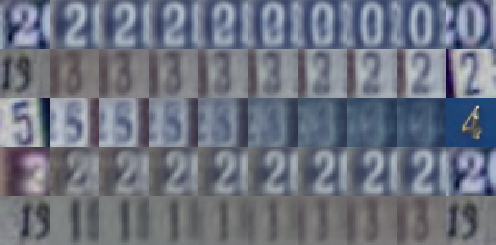}
    }
    \subfloat[\our{}]{
        \includegraphics[width=.24\linewidth]{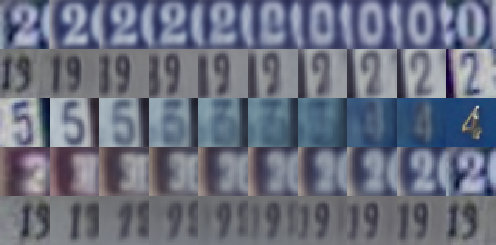}
    }
    \newline
    \subfloat[ADGM]{
        \includegraphics[width=.49\linewidth]{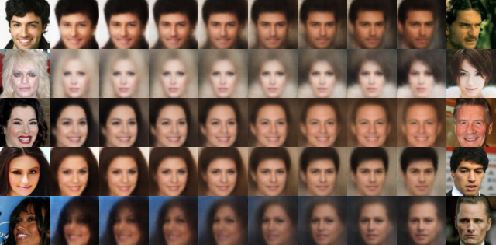}
    }
    \subfloat[\our{}]{
        \includegraphics[width=.49\linewidth]{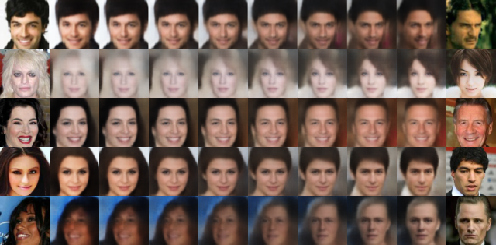}
    }
    \captionof{figure}{Class interpolations trained on various datasets. The first and the last columns are the real samples.}
    \label{fig:inter}
\end{figure*}

To provide an additional quantitative assessment of sample quality, we calculate the Fréchet Inception Distance (FID) \cite{heusel2017gans}, which is a standard score for generative models. To calculate FIDs, we used 10000 images sampled from the target distribution in the latent space. To better understand the connection between the generativity, semi-supervision, and a mixture of Gaussians as a target distribution, we also show FIDs for an unsupervised variant of \our{} (i.e. for $\beta=0$) and for a CWAE with a single spherical Gaussian as a target distribution (i.e. for $\beta=0$ and $K=1$). Moreover, to show the dependence of \our{} on the type of architecture, for SVHN we also report FIDs for a convolutional neural network. It is evident from Table \ref{tab:fid_scores} that the FIDs of \our{} and ADGM are comparable across datasets, which is slightly surprising when one looks at the samples generated by these models. The use of a convolutional architecture allows to significantly improve the performance of \our{} on SVHN. It is interesting that the FID of CWAE is better than both versions of \our{} on SVHN. It suggests that deep generative models can be trained more easily on unimodal target distributions than on a more complex mixture of Gaussians. In this case, we need to pay an additional cost (i.e., slightly lower quality of samples) for building a generative model for each class. 
Nevertheless, this observation does not hold for CelebA and MNIST, where the information of class labels allows to build better generative models for the entire data.


Visual inspection of the results suggests that the components in each model correspond to the correct classes (examples in each row of the figures have the same class). There are only few errors (e.g., the digits "2" and "3" in the second and sixth row of Fig.~\ref{fig:our-mnist} or the digit "5" in the ninth row of Fig.~\ref{fig:our-svhn}). Notably, similar errors occur also in M1+M2 (e.g., digit "6" in the first row of Fig.~\ref{fig:vae-svhn}) and ADGM (e.g., digit "8" in the last row of Fig.~\ref{fig:adgm-svhn}) despite the fact that these methods require explicit discrete latent variables. To get deeper insight into this aspect, we measured the classification accuracy on test sets. More precisely, we encoded each test image using an encoder network and classify it using~\eqref{eq:postG} in the latent space. Table~\ref{tab:accuracy} shows that \our{} gives better results than competitive methods on harder datasets such as SVHN and CelebA, especially when using a convolutional architecture. This is consistent with our previous observation that \our{} obtained higher quality of samples on these sets.

\subsection{Interpolation} \label{sec:interpolation}

ADGM and M1+M2
use a single Gaussian as a target distribution for the style of all classes, but add an additional discrete distribution to identify the class. By one-hot encoding this discrete class variable, interpolation between classes is possible in principle; e.g., by changing the class indicator from $(1,0,0,\dots)$ to $(0,1,0,\dots)$ via $(0.5,0.5,0,\dots)$. However, since $(0.5,0.5,0,\dots)$ has never appeared during training, these methods produce realistically-looking interpolations between samples in the same class, but have problems with smooth interpolation between samples in different classes. In contrast, \our{} uses a single continuous latent space in which the discrete class label is implicit in the identity of the mode of a potentially multimodal target distribution (e.g., Gaussian component in the case of a mixture of Gaussians). In other words, the samples from all classes are placed in the same continuous space. Ensuring that the different mixture components are not too far apart -- thus ensuring that also the space between component means is populated by training data -- allows us to obtain realistic interpolations.

To confirm this claim experimentally, we randomly pick a pair of images and map them to the latent space. Next, we generate linear interpolations between these encoded points and decode these interpolations using a decoder network. To obtain analogical effect for ADGM, we interpolate both the latent codes and the one-hot representations of classes.

\begin{figure*}[t]
\centering
    \centering
    \subfloat[ADGM]{
        \includegraphics[width=.24\linewidth]{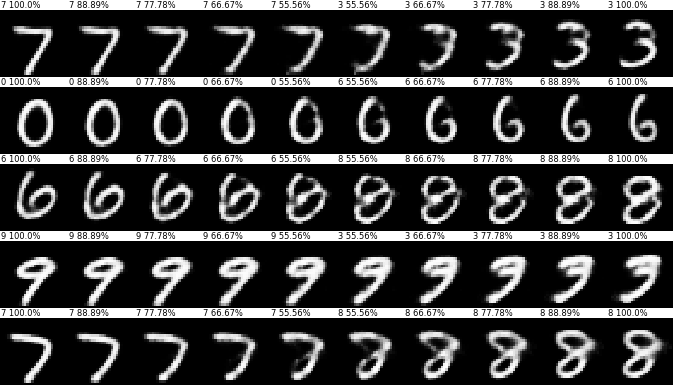}
    }
    \subfloat[\our{}]{
        \includegraphics[width=.24\linewidth]{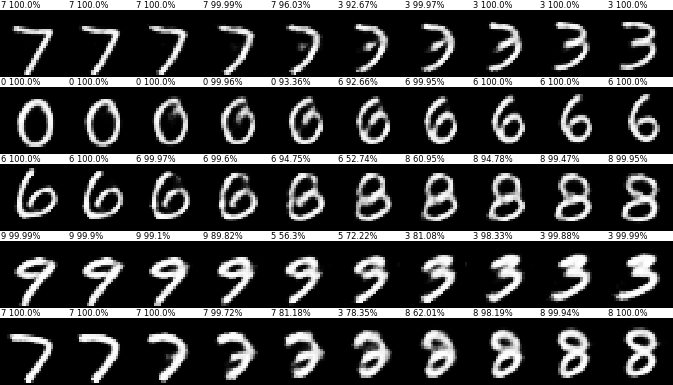}
    }
    \subfloat[ADGM]{
        \includegraphics[width=.24\linewidth]{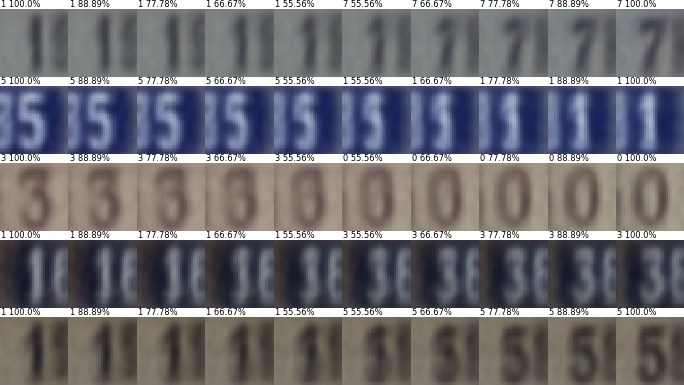}
    }
    \subfloat[\our{}]{
        \includegraphics[width=.24\linewidth]{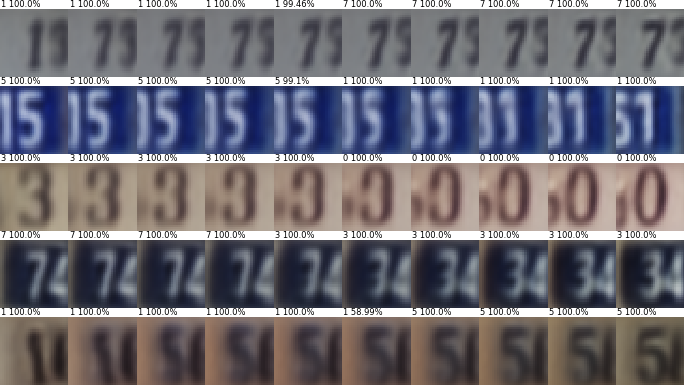}
    }
    \newline
    \subfloat[ADGM]{
        \includegraphics[width=.49\linewidth]{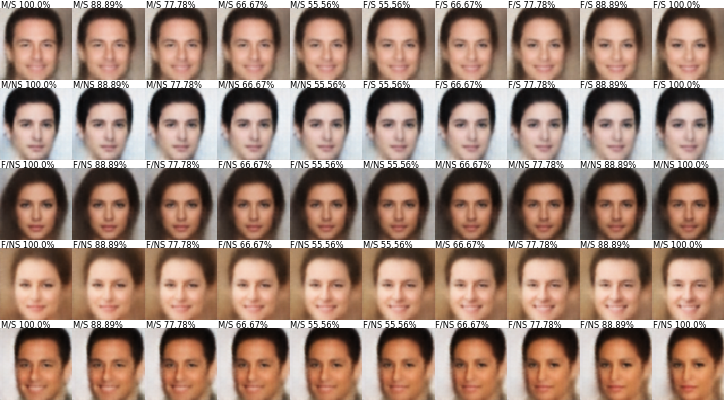}
    }
    \subfloat[\our{}]{
        \includegraphics[width=.49\linewidth]{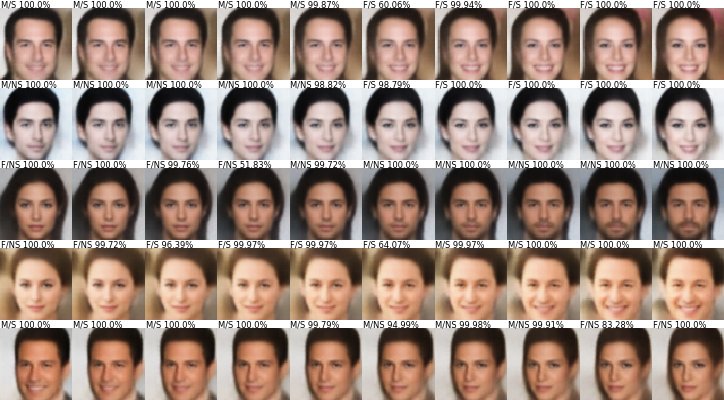}
    }
    \captionof{figure}{Smooth style transfer between classes.}
    \label{fig:classwise}
\end{figure*}

The results of ADGM and \our{} are presented in the Fig.~\ref{fig:inter} (we omit M1+M2 interpolations because of their poor quality). It can be seen that interpolations produced by \our{} have better quality than the ones obtained using ADGM. It is evident that many interpolations generated by ADGM are outside of the true data density (e.g. images in the sixth column of the first, second, fourth and fifth rows in the case of MNIST). For SVHN and CelebA, interpolations are often blurry or unrealistic (second row of CelebA and fifth row for SVHN). Note that the interpolations produced by \our{} can also pass through the regions of low density, which lie between Gaussian components (e.g. fourth row in the case of MNIST). This negative effect could be however partially fixed using more sophisticated interpolations than the linear one \cite{lesniak2018distribution, struski2019interpolation}, which is not fully justified for this type of distribution. Nevertheless, we can observe very interesting geometrical properties of the class structure obtained by \our{}: The interpolation between "6" and "9" (first row) leads to generating a digit "4" (first row). This suggests that the class of digits "4" is localized between classes of digits "6" and "9". Similar effect can be observed for the second row, where the model passes through the class of digits "3".

We support above claims by calculating FID scores for images generated from interpolations. Specifically, we took two random images from the test data, encoded them into the latent space, randomly picked a point on the line connecting the encoded images and decoded it. Table~\ref{tab:interpolation_fids} shows the FIDs for 10000 such decoded samples when compared against the real data. It is evident that FIDs of \our{} are significantly better than the ones obtained by competitive methods.

\subsection{Style Transfer and Class Intensity} \label{sec:intensity}

Semi-supervised generative models based on the VAE generate images by decoding a data point sampled from the target distribution and a one-hot vector describing the class label. If we flip the class vector, but keep the latent code unchanged, then the model returns an image with similar style, but from a different class. In other words, these models disentangle class and style information in latent space. In this experiment, we investigate whether a comparable style transfer can be achieved for our model.

In the case of \our{}, we cannot simply change the class label, because we do not have an individual vector describing the class. However, we can move a latent code generated by one Gaussian component to the corresponding location w.r.t.\ the mode of another Gaussian component. In other words, we move a latent code $z$ by a vector between the mean of target Gaussian $N(\mu_t, I)$ and the mean of the initial Gaussian $N(\mu_s, I)$ from which it was generated, i.e.:
\begin{equation} \label{eq:shift}
    \hat{z} = z + (\mu_t - \mu_s)
\end{equation}
In addition, we can interpolate between $z$ and $\hat{z}$ to change the class label gradually. Moving along this line in latent space is more natural than moving along a corresponding line between two one-hot encoded class vectors, as it would be necessary for semi-supervised VAE models. In the continuous latent space, it is natural to assume that close points correspond to similar images; for one-hot encoded class vectors or a discrete latent space, notions of closeness and similarity are inadequate.

Based on the above discussion, we encode an image into the latent space and gradually change its class label in a direction to the randomly selected class according to \eqref{eq:shift}. Sample results of this operation for \our{} and analogous results for ADGM are presented in Fig.~\ref{fig:classwise}. Although the style transfer leads to different images for \our{} and for ADGM (see right-most images in Fig.~\ref{fig:classwise}), it can be seen that the style is indeed maintained for \our{}. In addition to that, the images obtained via interpolation are more natural for \our{}. As mentioned, this is because ADGM encodes only discrete class labels while \our{} treats it as continuous variable.

\begin{figure}[t]
    \includegraphics[width=1.0\linewidth]{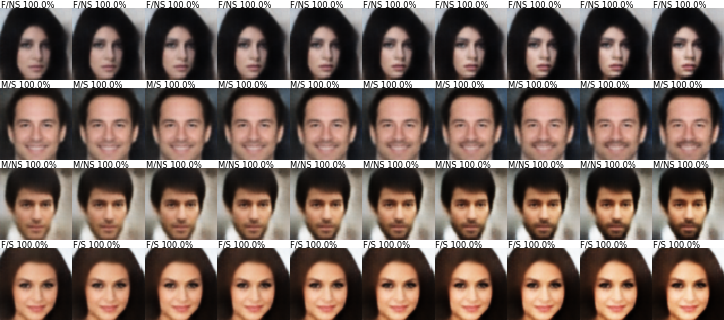}
    \caption{Changing intensity of the label. The samples in the first row are moving from the anti-target classes, which is "not male/smiling" for the first three rows and "male/smiling" for the fourth row.
    \label{fig:gmcwae_class_extra}}
\end{figure}

An interesting property of our model is that we can not only change the class of the presented data point, but also increase or decrease the intensity of certain stylistic aspects. We can achieve this property by moving the latent representation $z$ of a data point away from a class which lacks the considered stylistic aspect. More precisely, let $z$ be a latent code of an image that belongs to the initial Gaussian $N(\mu_s, I)$. We can move $z$ away from an anti-target Gaussian $N(\mu_t, I)$ by setting $\hat{z} = z + \alpha (\mu_s - \mu_t)$
for some $\alpha>0$. For example, by using an image of a male face and then moving it away from the Gaussian representing the `not male` class, we can add more masculine features to the picture. Examples of this operation are presented in Fig. \ref{fig:gmcwae_class_extra} for $\alpha = 0.5$. In the first three rows, the samples are moving away from the "not male/smiling" class, while in the fourth row, it is moving away from the "male/smiling" class. It can be observed that the man in the third row gains more masculine feature, while the woman in the fourth row gains more feminine features. This suggests that the presence of a beard is recognized as a typical male attribute and was encoded far from Gaussian components connected with female images.

\begin{figure}[t]
    \includegraphics[width=0.49\linewidth]{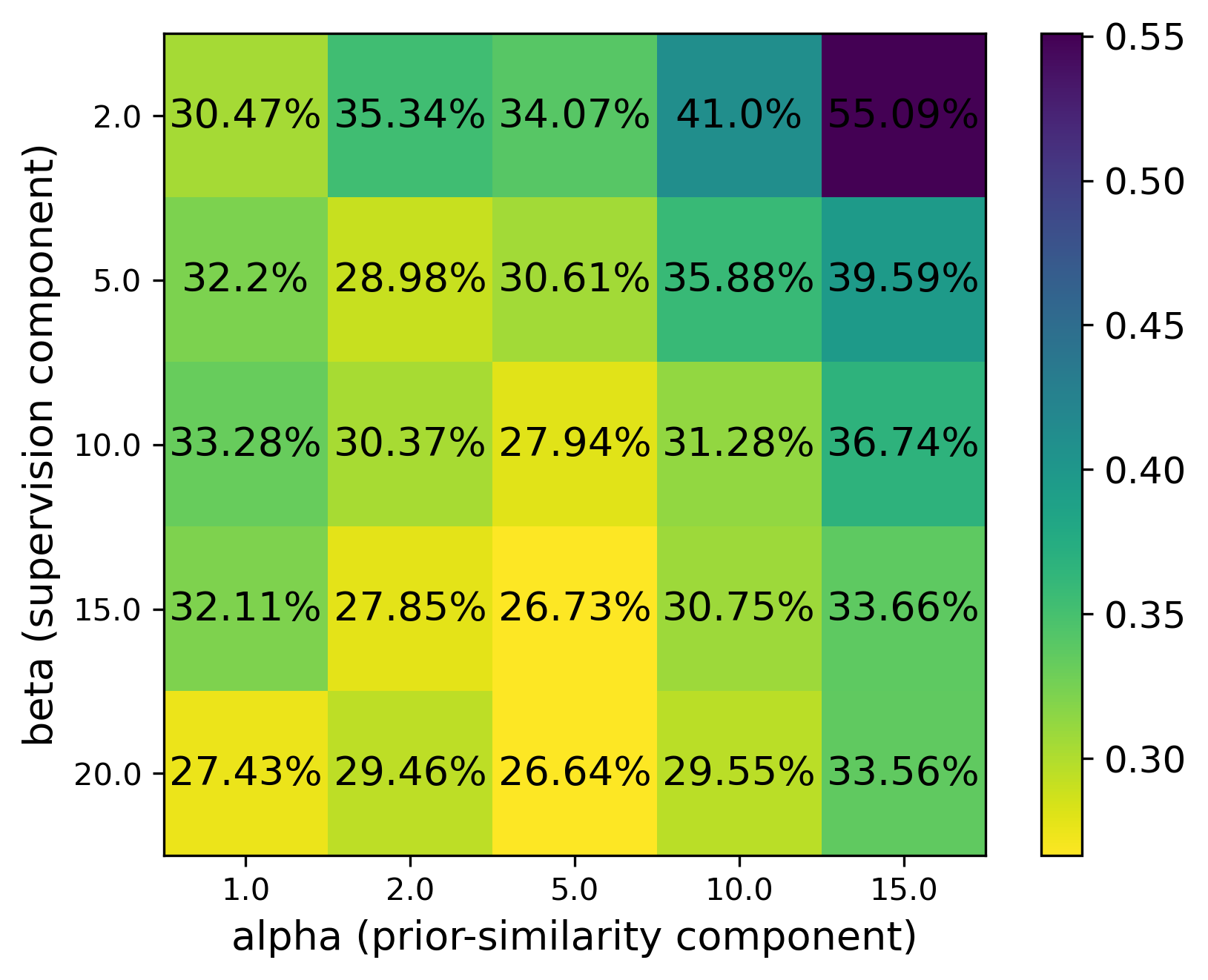}
    \includegraphics[width=0.49\linewidth]{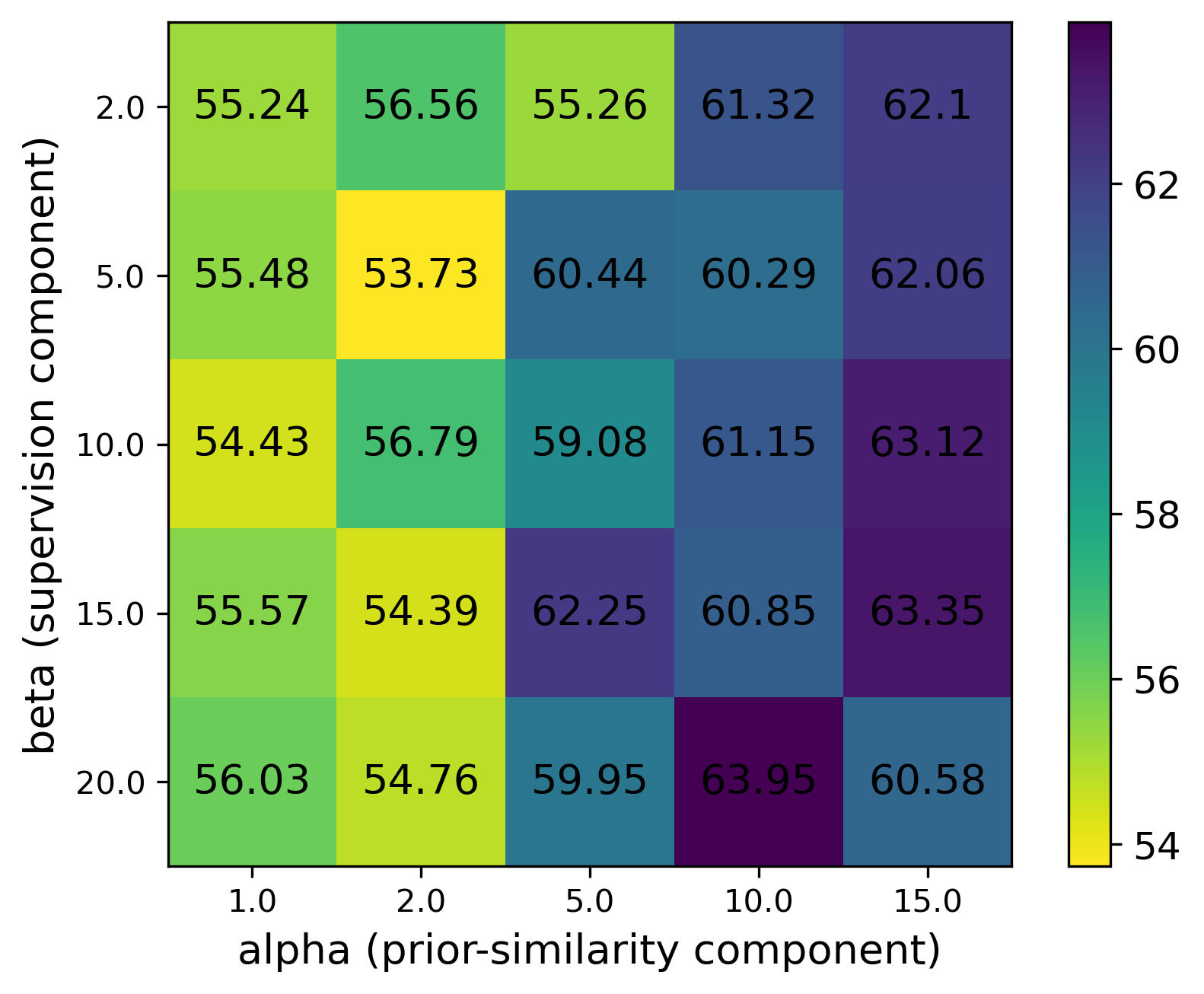}
    \caption{Hyperparemeter sensitivity analysis, considering test accuracies (left) and FID scores (right) for SVHN.
    \label{fig:hyperparam_analysis}}
\end{figure}

\begin{figure}[t]
    \includegraphics[width=0.49\linewidth]{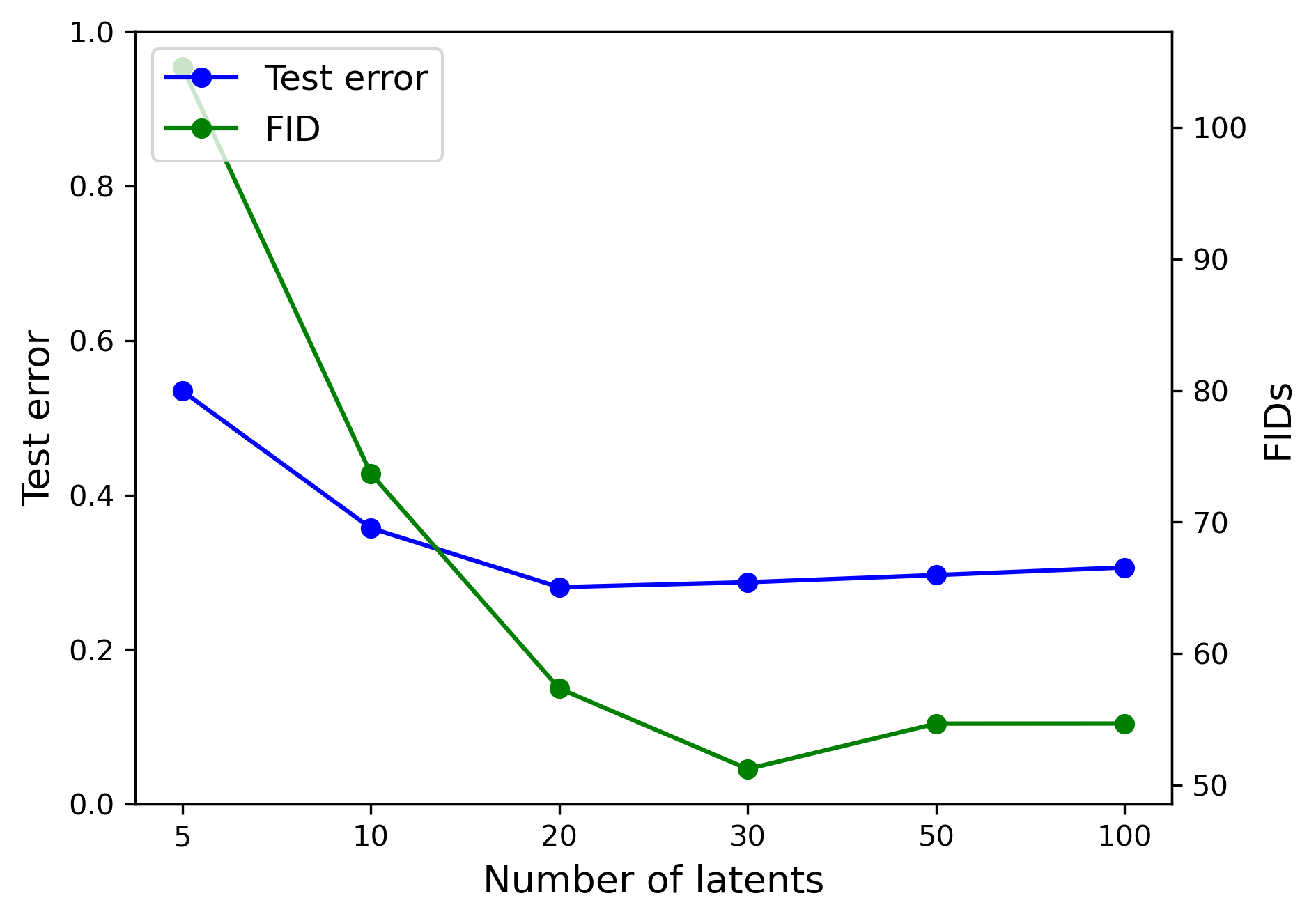}
    \includegraphics[width=0.49\linewidth]{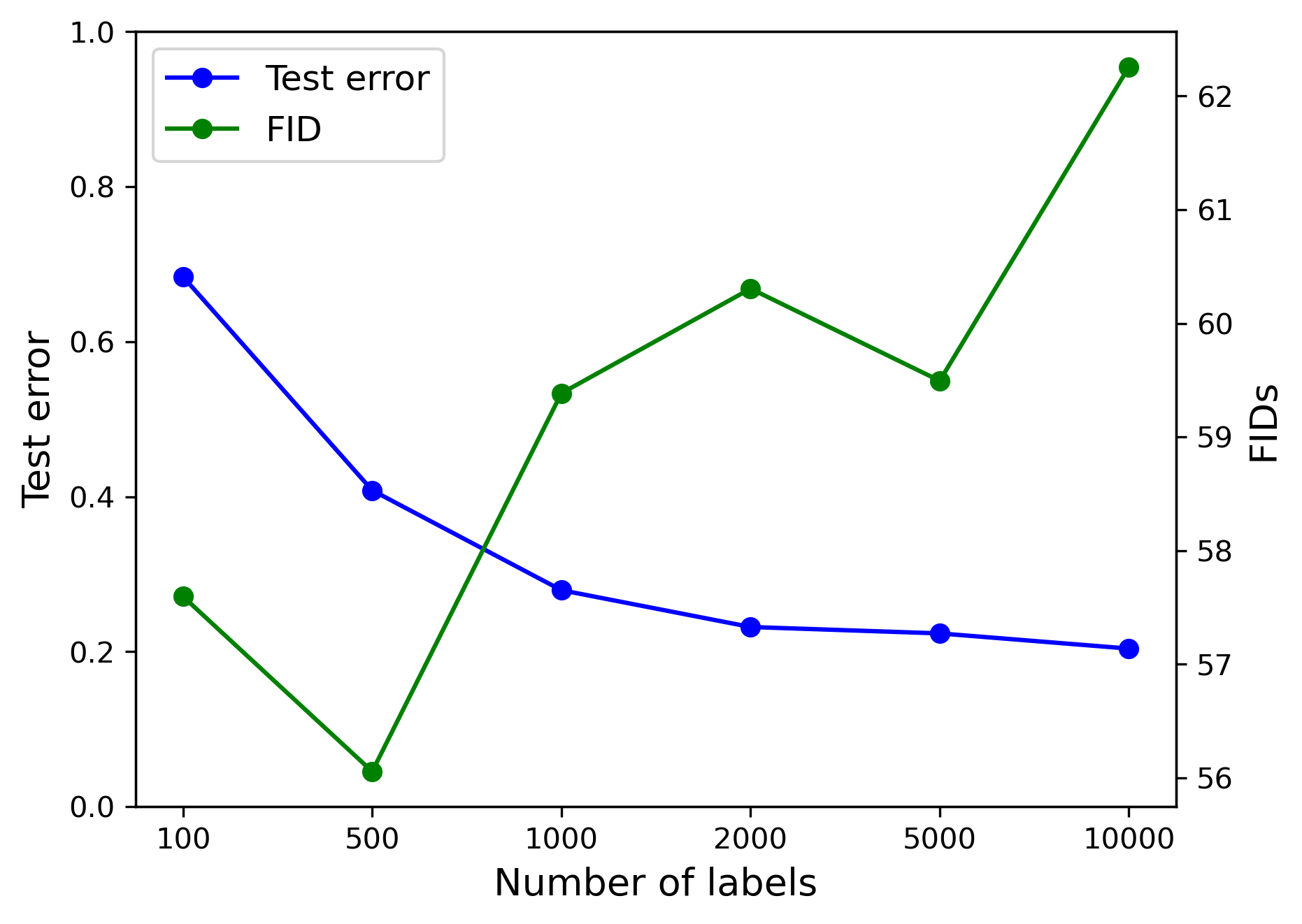}
    \caption{\mw{Influence of the number of latent dimensions (left) and labeled examples (right) on test accuracy and FID scores.}
    \label{fig:labels_dim}}
\end{figure}





\subsection{Sensitivity Analyses} \label{sec:sensitivity}

We finally analyze the sensitivity of \our{} to its parameters. For the sake of brevity we only present results for the fully connected architecture trained on SVHN.

First, we examine the sensitivity to the selection of hyperparameters. Fig.~\ref{fig:hyperparam_analysis} shows test accuracy and FIDs obtained for different values of $\alpha$ (parameter for Cramer-Wold loss) and $\beta$ (parameter for cross-entropy loss). 
Visual inspection suggests that the loss component representing the reconstruction error is crucial for both classification and generativity. The reconstruction error component has no hyperparameter, i.e., its relative weight increases if $\alpha$ and $\beta$ are both small. Indeed, increasing either $\alpha$ or $\beta$ without increasing the capacity of the model (number of layers and their width, etc) increases FIDs and decreases test accuracies, because the network no longer tries to distinguish important details of the examples. While $\beta$ is positively correlated with test accuracy its influence on FIDs seems weak.
More generally, judging from the range of accuracies and FIDs in Fig.~\ref{fig:hyperparam_analysis} (most accuracies and FIDs range between 65\% and 74\% and between 54 and 62, respectively), we conclude that \our{} is comparably stable w.r.t.\ the hyperparameters, which obviates the need for fine-tuning.

We next analyze the influence of the number of latent dimensions on the performance. The results presented in Fig. \ref{fig:labels_dim} (left) demonstrate that a too small latent space creates a bottleneck that prevents the model from separating examples from different classes and from generating high-quality samples. Introducing too many dimensions in the latent space leads to a small but noticeable decrease of performance. In summary, however, the sensitivity of \our{} on the number of latent dimensions is weak.

Finally, we examine the dependence on the number of labels available in the training set. As expected, Fig.~\ref{fig:labels_dim} (right) shows that a higher number of labeled data leads to better classification accuracy. In contrast, increasing the number of labeled samples seems to have a small negative effect on the quality of generated samples (measured by FID scores). As the network has to fit more labeled data, a smaller portion of its capacity can be used to reduce other loss components. This shows that generative models are capable of producing realistic images without any information from class labels. This is consistent with the results in Table~\ref{tab:fid_scores}, where unsupervised \our{} achieved better FIDs than its supervised version. 

\section{Conclusion}

We constructed \our{}, a semi-supervised generative model, which combines a WAE-MMD framework with a mixture of Gaussians as a target distribution. By connecting each component with a specific class, \our{} is capable of generating class representatives of high quality. Since a discrete class label is implicitly encoded in the same latent space as the style variable, we can interpolate between samples from different classes, which was not possible in existing semi-supervised generative models. In particular, we can change the class labels gradually while preserving the style of a given sample. Moreover, \our{} allows for increasing (decreasing) the influence of a given class by moving this example in the (opposite) direction of the corresponding Gaussian component. For example, moving a sample away from the ``female'' components in latent space resulted in growing a beard in male celebrities.


Due to the use of a simple auto-encoder architecture from a classical WAE model, \our{} scales better with the complexity of the dataset than other semi-supervised generative approaches. In particular, \our{} requires fewer dimensions in the latent space (20 dimensions for SVHN) than ADGM (300 dimensions for SVHN), which is composed on six neural networks. Making use of Cramer-Wold distance as MMD penalty, we do not need to sample during training, which saves computational resources and increases model stability as compared to, e.g. ADGM which requires sampling. 
Finally, \our{} can be more easily adapted to different (e.g. convolutional) network architectures than the previous semi-supervised models.

\appendix
\section{Architectural Details for Experiments with CelebA}
For \our{}, the encoder $\E$ consists of two convolutional layers with 32 filters, two convolutional layers with 64 filters, and three fully connected layers of size $786-786-32$\newmw{, where $D=32$ is the latent space dimension}. The decoder $\D$ consists of three fully connected layers of size $786-786-1024$, one transposed convolutional layer with 64 filters, two transposed convolutional layers with 32 filters, and a final transposed convolutional layer with 3 filters.

For M1, the encoder $\E$ consists of two convolutional layers with 32 filters, two convolutional layers with 64 filters, and two fully connected layers of size $300-50$\newmw{, where $D=50$ is the latent space dimension}. The decoder $\D$ consists of two fully connected layers of size $300-1024$, one transposed convolutional layer with 64 filters, two transposed convolutional layers with 32 filters, and a final transposed convolutional layer with 3 filters. The latent space has dimension $D=300$. For M2, encoder and decoder are fully connected with sizes $300$ and $300-300$, respectively.

For ADGM, the encoder $\E$ consists of two convolutional layers with 32 filters and two convolutional layers with 64 filters. The result is a flattened vector. After that every edge in the probabilistic graph of this model is implemented by a neural network with two fully connected layers of size $500-500$\newmw{, with the latent space dimension $D=300$}. Also the decoder $\D$ implements every edge in the probabilistic graph by a neural network with two fully connected layers of size $500-500$. The resulting image is further processed by a transposed convolutional layer with 64 filters, two transposed convolutional layer with 32 filters, and a final transposed convolutional layer with 3 filters.

All networks are implemented using ReLU activation functions and all convolution filters are of size 4x4 and stride 2x2.

\bibliographystyle{IEEEtran}
\bibliography{IEEEabrv,ref}

\end{document}